\documentclass[runningheads,a4paper]{llncs}

\usepackage{amssymb}
\usepackage{graphicx}
\usepackage{bm,amsmath} 
\usepackage{color}

\usepackage{url}
\urldef{\mailsa}\path|{alfred.hofmann, ursula.barth, ingrid.haas, frank.holzwarth,|
\urldef{\mailsb}\path|anna.kramer, leonie.kunz, christine.reiss, nicole.sator,|
\urldef{\mailsc}\path|erika.siebert-cole, peter.strasser, lncs}@springer.com|

\begin{document}

\mainmatter  

\title{Learning Interpretable Anatomical Features Through Deep Generative Models:\\ Application to Cardiac Remodeling}

\titlerunning{Deep Learning of Interpretable Anatomical Features}

%
%
\author{Carlo Biffi, Declan P. O'Regan, Daniel Rueckert}
\authorrunning{Biffi et al.}

\author{Carlo Biffi\inst{1,2} \and
Ozan Oktay\inst{1} \and
Giacomo Tarroni\inst{1} \and
Wenjia Bai\inst{1} \and
Antonio De Marvao\inst{2} \and
Georgia Doumou\inst{2} \and
Martin Rajchl\inst{1} \and
Reem Bedair\inst{2} \and
Sanjay Prasad\inst{3} \and
Stuart Cook\inst{3,4} \and
Declan O'Regan\inst{2} \and
Daniel Rueckert\inst{1}}
\authorrunning{Biffi et al.}
%
\institute{Biomedical Image Analysis Group, Imperial College London, London, UK \and
MRC London Clinical Sciences Centre, Imperial College London, UK \and
National Heart \& Lung Institute, Imperial College London, London, UK \and
Duke-NUS Graduate Medical School, Singapore}

\footnotetext[1]{The research was supported by grants from the British Heart Foundation (NH/17/1/32725, RE/13/4/30184).}

\toctitle{Lecture Notes in Computer Science}
\tocauthor{Authors' Instructions}
\maketitle

\begin{abstract}
Alterations in the geometry and function of the heart define well-established causes of cardiovascular disease. However, current approaches to the diagnosis of cardiovascular diseases often rely on subjective human assessment as well as manual analysis of medical images. Both factors limit the sensitivity in quantifying complex structural and functional phenotypes. Deep learning approaches have recently achieved success for tasks such as classification or segmentation of medical images, but lack interpretability in the feature extraction and decision processes, limiting their value in clinical diagnosis. In this work, we propose a 3D convolutional generative model for automatic classification of images from patients with cardiac diseases associated with structural remodeling. The model leverages interpretable task-specific anatomic patterns learned from 3D segmentations. It further allows to visualise and quantify the learned pathology-specific remodeling patterns in the original input space of the images. This approach yields high accuracy in the categorization of healthy and hypertrophic cardiomyopathy subjects when tested on unseen MR images from our own multi-centre dataset (100\%) as well on the ACDC MICCAI 2017 dataset (90\%). We believe that the proposed deep learning approach is a promising step towards the development of interpretable classifiers for the medical imaging domain, which may help clinicians to improve diagnostic accuracy and enhance patient risk-stratification.

\end{abstract}

\section{Introduction}
Alterations in the geometry and function of the heart (remodeling) are used as criteria to diagnose and classify cardiovascular diseases as well as risk-stratify individual patients \cite{Cohn}. For instance, hypertrophic cardiomyopathy (HCM), a leading cause of sudden death in adults \cite{Yancy}, is an inherited disease of the heart muscle which manifests clinically with unexplained left ventricular (LV) hypertrophy and can occur in many different patterns that are not readily quantifiable \cite{Captur}. Cardiovascular magnetic resonance (CMR) has become the gold-standard imaging technique for quantitative assessment in the diagnosis and risk-stratification of cardiomyopathy \cite{Elliott}. However, image interpretation is often dependent on both clinical expertise and effective diagnostic criteria, making automated data-driven approaches appealing for patient classification - especially as conventional manual analysis is not sensitive to the complex phenotypic manifestations of inherited heart disease. In recent years, large population cohorts have been recruited, such as the UK Biobank study with cardiac imaging in up 100,000 participants \cite{Petersen}, requiring new approaches to high-throughput analysis. Learning-based approaches that can capture complex phenotypic variation could offer an objective data-driven means of disease classification without human intervention. Indeed, early work has shown the potential of machine learning algorithms in distinguishing benign from pathological hypertrophy from multiple manually-derived cardiac parameters \cite{Nerula}.

Deep learning approaches have recently achieved outstanding results in the field of medical imaging due to their ability to learn complex non-linear functions, but they lack interpretability in the feature extraction and decision processes, limiting their clinical value. In this work, we propose a variational autoencoder (VAE) model \cite{Kingma} based on 3D convolutional layers, which is employed for classification of cardiac diseases associated with structural remodeling. This generative model enables us to visualise and leverage interpretable task-specific anatomical patterns learned from the segmentation data. The performance of the proposed approach is evaluated for the classification of healthy volunteers (HVols) and HCM subjects on our own dataset multi-centre cohort and on the ACDC MICCAI 2017 challenge dataset. This work makes two major contributions. First, we introduce a deep learning architecture which can discriminate between different clinical conditions through task-specific interpretable features, making the classification decision process transparent. Second, we develop a method to visualise and quantify the learned pathology-specific remodeling in the original space of the images, providing a data-driven method to study complex phenotypic manifestations.

\subsection{Related work}
Given the high dimensionality of medical images, a popular approach in the literature is to analyze them by constructing statistical shape models of the heart using finite elements models or segmentation and co-registration algorithms to derive subject-specific meshes \cite{Pau,Bai}. Similar to brain image analysis, principal component analysis (PCA) is then subsequently performed on these point distribution models to learn their main modes of deformation. These modes are then employed in the discrimination of distinct groups of subjects by their shape differences or to identify the ones mostly associated with diseases \cite{Remme,Ardekani,Zhang}. However, PCA shape components do not define the features that differentiate between disease classes. For this purpose, approaches that search for new axes of variation that are clinically-meaningful have been proposed \cite{Suinesiaputra,GIGA}. Relevant to this work, in the brain imaging domain, Shakeri et al. \cite{Shakeri} employed a VAE model based on two fully connected layers to learn a low-dimensional representation of co-registered hippocampal meshes, which is later employed in a multi-layer perceptron to classify patients with Alzheimer disease. By contrast, the proposed method can exploit a deep convolutional neural network architecture directly on the segmentation maps to learn a discriminative latent space in an end-to-end fashion.

\section{Material and Methods}

\subsection{Datasets}
A multi-centre cohort consisting of 686 HCMs patients (57 $\pm$ 14 years, 27\% women, 77\% Caucasian, HCM diagnosed using standard clinical criteria) and 679 healthy volunteers (40.6 $\pm$ 12.8 years, 55\% women, 69\% Caucasian) was considered for this work. Participants underwent CMR at 1.5-T on Siemens (Erlangen, Germany) or Philips (Best, Netherlands) systems. Cine images were acquired with a balanced steady-state free-precession sequence and included a stack of images in the left ventricular short axis plane (voxel size 2.1x1.3x7 mm$^3$, repetition time/echo time of 3.2/1.6 ms, and flip angle of $60^\circ$). End-diastolic (ED) and end-systolic (ES) phases were segmented using a previously published and extensively validated cardiac multi-atlas segmentation framework \cite{Bai}. As a first preprocessing step, we improved the quality of the 2D stacks segmentation by a multi-atlas-aided upsampling scheme. For each segmentation, twenty manually-annotated high-resolution atlases at ED and ES were warped to its space using a landmark-based rigid registration. Then a free-form non-rigid registration with a sparse set of control points was applied (nearest-neighbor interpolation) \cite{Rueckert} and fused with a majority voting consensus. In a second step, we aligned all the quality-enhanced segmentations onto the same reference space at ED by means of landmark-based and subsequent intensity-based rigid registration to remove pose variations. After extracting the LV myocardium label, we cropped and padded each segmentation to $[x=80, y=80, z=80, t=1]$ dimension using a bounding box centered at the LV’s ED myocardium. The latter operation guarantees shapes to maintain their alignment after cropping. Finally, all the segmentations underwent manual quality control in order to discard scans with strong inter-slice motion or insufficient LV coverage. As an additional testing dataset, 20 HVols and 20 HCMs from the ACDC MICCAI’17 challenge training dataset, consisting of 2D MR image sequences which are annotated at ED and ES phases by a clinical expert, were pre-processed using the same pipeline explained above.

\subsection{Deep Generative Model}
\begin{figure}
\centering
\includegraphics[width=\textwidth]{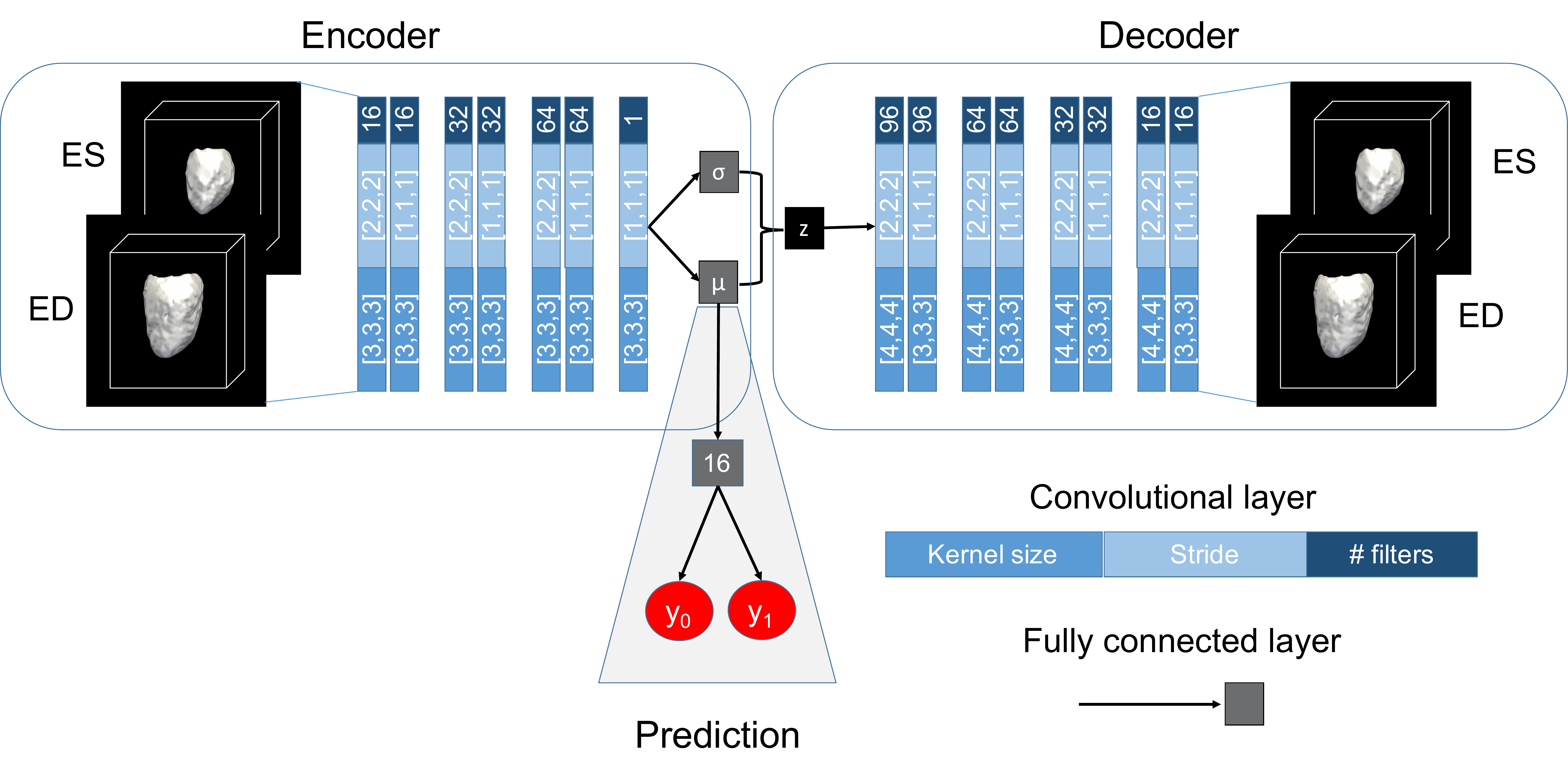}
\caption{Generative model architecture. Registered LV segmentations at ED and ES phases are mapped to a low-dimensional latent space. Each latent dimension is forced to be normally distributed with mean $\bm{\mu}$ and standard deviation $\bm{\sigma}$. A decoder network is then used to reconstruct the input segmentation from a low-dimensional vector $\bm{z}$ sampled from the learned latent distribution (training) or the $\bm{\mu}$ vector (testing). The $\bm{\mu}$ latent representation is used as input of a MLP to predict disease status.}
\label{fig:arc}
\end{figure}
\subsubsection*{Architecture} A schematic representation of the proposed architecture is shown in Fig. \ref{fig:arc}. The network input $\bf{X}$ consists of subjects' 3D LV myocardial segmentations at ED and ES phases presented as a two-channel input. A 3D convolutional VAE is employed to learn a $d$-dimensional probability distribution representing the input segmentations $\bf{X}$ in a latent space through an encoder network. In this work, this latent distribution is parametrized as $d$-dimensional normal distribution $\mathcal{N}(\mu_i,\sigma_i)$ with mean $\mu_i$ and standard deviation $\sigma_i$, $i=1,\dots,d$. During training, a decoder network learns to reconstruct approximations of the input segmentations $\bf{X}$, which are denoted as $\bf{\hat{X}}$, by sampling vectors $\bf{z}$ from the learned latent $d$-dimensional manifold $\mathcal{Z}$, $\bf{z} \in \mathcal{Z} = \mathcal{N}(\bm{\mu},\bm{\sigma})$. Simultaneously, a discriminative network (which is referred to as prediction network in the context of the paper) constructed with a multilayer perceptron (MLP) is connected to the mean vector $\bm{\mu}$ and trained to discriminate between HVols and HCMs subjects. This architecture is trained end-to-end with a loss function of the form $\mathcal{L} = \mathcal{L}_{rec} + \alpha \mathcal{L}_{KL} + \beta \mathcal{L}_{MLP}$. $\mathcal{L}_{rec}$ is the reconstruction loss and it was implemented as a Sorensen-Dice loss between the input segmentations $\bf{X}$ and their reconstruction $\bf{\hat{X}}$. $\mathcal{L}_{KL}$ is the Kullback-Leibler divergence loss forcing $\mathcal{N}(\bm{\mu},\bm{\sigma})$ to be as close as possible to its prior distribution $\mathcal{N}(\bf{0},\bf{1})$. $\mathcal{L}_{MLP}$ is the cross-entropy loss for the MLP classification task. The latent space dimension was fixed to $d=64$. At test time, each input segmentation is reconstructed by passing the predicted $\bm{\mu}$ to $\bf{z}$ (without sampling from the latent space), while the classification is performed as in training time. 

\subsubsection*{Interpreting Learned Features via Navigation in the Latent Space} Our generative model architecture allows for visualization of the features learned by the network in the original segmentation space. For this, the weights learned by the MLP can be exploited to compute the partial derivative of the disease class label C ($y_{C}$) w.r.t. to the latent space representation $\bm{\mu}$ of an input $X$, i.e. $\frac{\partial y_{C}}{\partial \mu_i}$, by backpropagating the gradient from the class label C to $\mu_i$ using chain-rule.
Given a randomly selected healthy shape, we can use the derived gradient to move the latent representation of a subject $\bm{\mu}$ along the direction of the latent code variability that maximises the probability of its classification to class C using an iterative algorithm. Starting with the mean latent representation $\bm{\mu}_0=\bm{\bar{\mu}}$ of a healthy shape we can iteratively update $\mu_{i}$ at each step $t$ accordingly to Eq. 1:
\begin{equation}
\mu_{i,t} = \mu_{i,t-1} + \lambda \; \frac{\partial y_{\tiny 1}}{\partial \mu_{i,t-1}}, \;\;\; \forall i=1,\dots d
\end{equation}
\noindent Here we use $\lambda = 0.1$. Finally, each latent representation $\bm{\mu}_t$ at each step $t$ can be decoded back to the segmentation space by passing it to $\bf{z}$, allowing for the visualization of the corresponding reconstructed segmentations $\bf{\hat{X}}$.



\section{Results}
Our dataset was split into training, evaluation and testing sets consisting of 537 (276 HVols, 261 HCMs), 150 (75 HVols, 75 HCMs) and 200 (100 HVols, 100 HCMs) subjects respectively. The model was developed in Tensorflow, and trained on a Nvidia Tesla K80 GPU using Adam Optimizer, learning rate of $10^{-4}$ and batch size of 16. After 96k iterations, the total validation loss function stopped improving and the training was stopped. No significant changes in the classification results were found by varying the loss parameters $\alpha$ and $\beta$, while $\alpha$ was set to 0.1 as this captured local shape variations without losing the generative model properties. All the 200 subjects in the testing dataset were correctly classified (100\% accuracy) by the trained prediction network. The model also correctly classified 36 out of the 40 ACDC MICCAI 2017 segmentations (90\% accuracy); of the 4 misclassified cases, 3 did not properly cover the whole LV, which might be the cause for the error. 

\begin{figure}[t!]
\centering
\includegraphics[width=\textwidth]{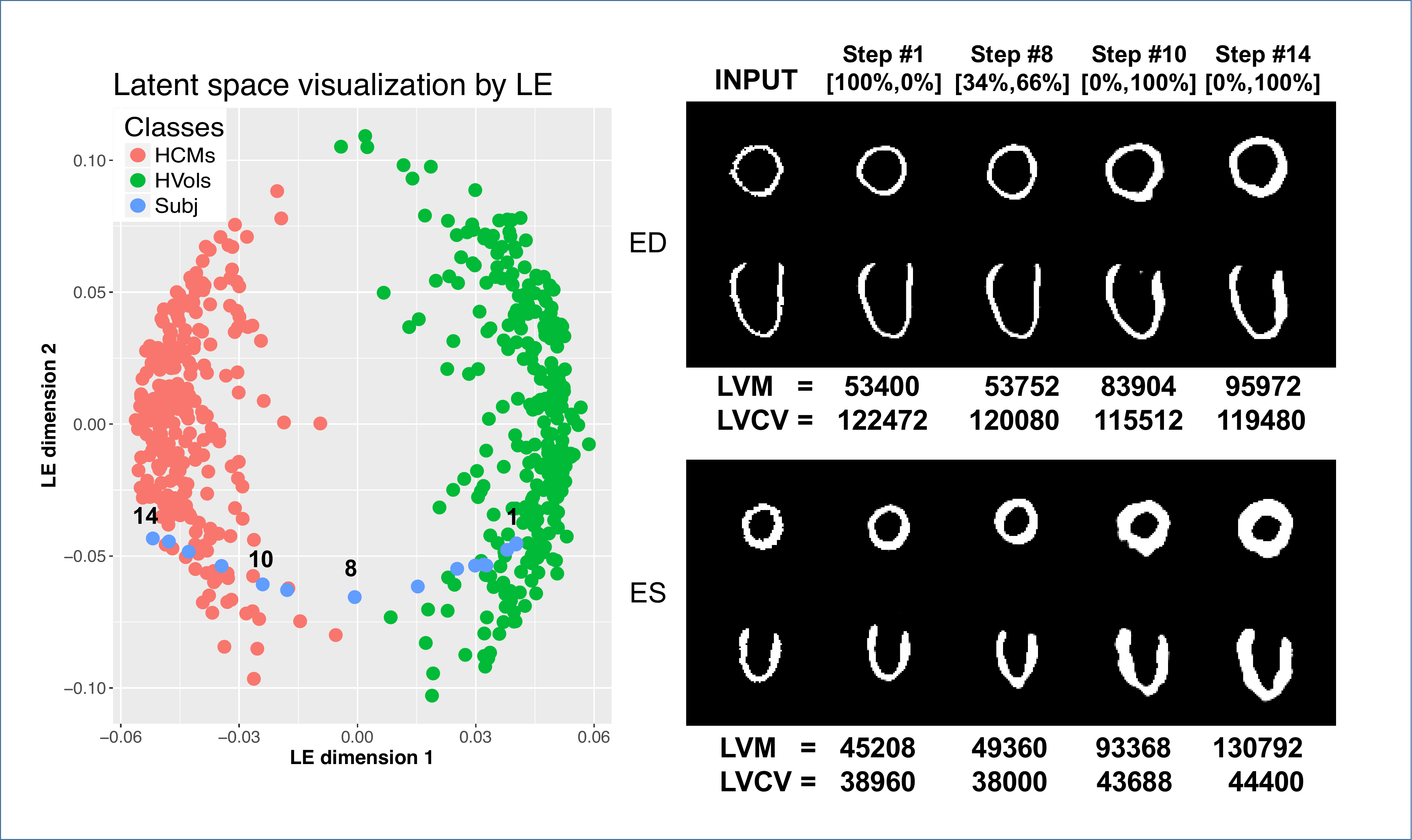}
\caption{On the left, Laplacian Eigenmaps (LE) bi-dimensional representation of the latent $\bm{\mu}$ of each subject in the training set (red and green dots) and of the $\bm{\mu}_t$ obtained through latent space navigation (light blue dots) for a random healthy shape. 
This latter is displayed on the right, together with the decoded segmentations corresponding to the sampled $\bm{\mu}_t$ reported on the left at 4 exemplary iterations. The probabilities of class HVOls and HCM, and the computed LVM and LVCV are also shown.}
\label{fig:mode}
\end{figure}

By employing the proposed method for latent space navigation, we deformed a randomly selected healthy segmentation from the training set towards the direction that maximizes its probability of being classified as HCM. On the right of Fig. \ref{fig:mode}, we report the original segmentations of the selected subject at ED and ES phases, their reconstruction from the VAE, and the reconstructed segmentations at four different iterations of the latent space navigation method. On the left of Fig. \ref{fig:mode}, the latent 64-dimensional representation $\bm{\mu}$ of the training set segmentations together with the latent representations $\bm{\mu}_t$ obtained at each iteration $t$ were reduced for visualization purposes to a bi-dimensional space using Laplacian Eigenmaps \cite{Belkin}. This technique allows to build a neighborhood graph of the latent representations that can be used to monitor the transformation (light blue points) of the segmentation under study from the HVol cluster to the HCM cluster. At each reported step, LV mass (LVM) from each segmentation was derived by computing the volume of the myocardial voxels. Moreover, a LV atlas segmentation having also labels for the LV cavity was non-rigidly registered to each segmentation to compute LV cavity volume (LVCV) by computing the volume of the blood pool voxels. Finally, for each iteration we also report the probabilities of being an HVol or HCM as computed by the prediction network. The learned deformations demonstrate a higher LVM and lower LVCV with an asymmetric increase in septal wall thickness in the geometric transition from HVol to HCM - which is the typical pattern of remodeling in this disease \cite{Desai}. At iteration 8, where the prediction network gives an indeterminate classification probability, LV geometry appears normal at ED but is thickened at ES suggesting that altered contractility may also be a discriminative feature.

\section{Discussion and Conclusion}
We present a deep generative model for automatic classification of heart pathologies associated with cardiac remodeling which leverages explainable task-specific anatomical features learned directly from 3D segmentations. The proposed architecture is specifically designed to enable the visualization and quantification of the learned features in the original segmentation space, making the classification decision process interpretable and potentially enabling quantification of disease severity. In this work we also propose a simple method that allows navigation in the low-dimensional manifold learned by the network, showing the potential clinical utility of the derived latent representation for tracking and scoring patients against a reference population. In the reported exemplar clinical application, the learned features achieved high accuracy in the discrimination of healthy subjects from HCM patients on our unseen testing dataset and on the ACDC MICCAI 17 dataset. 

The proposed architecture can be easily extended to other cardiac related clinical tasks by replacing the prediction network with survival, risk score or other clinical models implemented as neural networks. The proposed approach worked successfully on conventional MR acquisitions, showing its potential for using routinely acquired clinical MR imaging. Moreover, the use of segmentation masks could allow its application to a wider range of CMR images such as multi-site images acquired from different machines and using different imaging protocols. We acknowledge that our external testing dataset was small, in future work we plan to evaluate the proposed approach on a bigger unseen dataset from different centres and on various types of cardiomypathies. Further extensions will also include the integration of clinical variables to the latent space as well as the inclusion of other cardiac phases. 

The proposed approach is a promising step towards the development of interpretable deep learning classifiers for the medical imaging domain, which may assist clinicians to improve diagnosis and provide new insight into patient stratification. This general approach is not limited to the cardiac domain and can potentially be extended to other image analysis tasks where pathological shape change is prognostically relevant.

\end{document}